\documentclass[a4paper,fleqn]{cas-dc}

\usepackage[round,authoryear]{natbib}
\usepackage{amsmath}
\usepackage{soul}
\usepackage{bm}
\usepackage{color, xcolor}
\usepackage{caption}
\usepackage{subcaption}
\usepackage{algorithm}
\usepackage{algorithmic}
\usepackage{bbding}
\soulregister\cite7
\soulregister\citep7
\soulregister\ref7

\newcommand{\revise}[1]{{\color{black}#1}}

\def\tsc#1{\csdef{#1}{\textsc{\lowercase{#1}}\xspace}}
\tsc{WGM}
\tsc{QE}

\begin{document}
\let\WriteBookmarks\relax
\def\floatpagepagefraction{1}
\def\textpagefraction{.001}

\shorttitle{Y. Geng et~al./ Expert Systems with Applications}
\shortauthors{Y. Geng et~al.}

\title [mode = title]{Prompting Disentangled Embeddings for Knowledge Graph Completion with Pre-trained Language Model}



%

\author[1,2]{Yuxia Geng}[orcid=0000-0002-2461-2613]
\cormark[1]
\ead{geng_yx1@hdec.com,yuxia.geng@hdu.edu.cn}
\affiliation[a]{organization={Powerchina Huadong Engineering Corporation Limited},
            city={Hangzhou},
            postcode={311112}, 
            country={China}}

\affiliation[b]{organization={School of Computer Science, Hangzhou Dianzi University},
            city={Hangzhou},
            postcode={310018}, 
            country={China}}

\author[3]{Jiaoyan Chen}   
\ead{jiaoyan.chen@manchester.ac.uk}
\affiliation[c]{organization={Department of Computer Science, The University of Manchester},
            city={Manchester},
            postcode={M13 9PL}, 
            country={UK}}
            
\author[4]{Yuhang Zeng}
\ead{yuhang.zeng@hdu.edu.cn}
\affiliation[d]{organization={HDU-ITMO Joint Institute, Hangzhou Dianzi University},
            city={Hangzhou},
            postcode={310018}, 
            country={China}}
            
\author[5]{Zhuo Chen}[]
\ead{zhuo.chen@zju.edu.cn}
\affiliation[e]{organization={College of Computer Science and Technology, Zhejiang University},
            city={Hangzhou},
            postcode={310028}, 
            country={China}}

\author[6]{Wen Zhang}[]
\ead{zhang.wen@zju.edu.cn}
\affiliation[f]{organization={School of Software Technology, Zhejiang University},
            city={Ningbo},
            postcode={315048}, 
            country={China}}

\author[7]{Jeff Z. Pan}[]
\ead{j.z.pan@ed.ac.uk}
\affiliation[g]{organization={School of Informatics, The University of Edinburgh},
city={Edinburgh},
postcode={EH8 9AB},
country={UK}}

\author[2]{Yuxiang Wang}[]
\ead{lsswyx@hdu.edu.cn}

\author[2]{Xiaoliang Xu}[]
\ead{xxl@hdu.edu.cn}

\cortext[1]{Corresponding author.}



\begin{abstract}
Both graph structures and textual information play a critical role in Knowledge Graph Completion (KGC). With the success of Pre-trained Language Models (PLMs) such as BERT, they have been applied for text encoding for KGC. However, the current methods mostly prefer to fine-tune PLMs, leading to huge training costs and limited scalability to larger PLMs. In contrast, we propose to utilize prompts and perform KGC on a frozen PLM with only the prompts trained. Accordingly, we propose a new KGC method named PDKGC with two prompts --- a hard task prompt which is to adapt the KGC task to the PLM pre-training task of token prediction, and a disentangled structure prompt which learns disentangled graph representation so as to enable the PLM to combine more relevant structure knowledge with the text information. With the two prompts, PDKGC builds a textual predictor and a structural predictor, respectively, and their combination leads to more comprehensive entity prediction.
Solid evaluation on \revise{three} widely used KGC datasets has shown that PDKGC often outperforms the baselines including the state-of-the-art, and its components are all effective. Our codes and data are available at \href{https://github.com/genggengcss/PDKGC}{https://github.com/genggengcss/PDKGC}. 
\end{abstract}


\maketitle





\section{Introduction}
Knowledge Graphs (KGs)~\citep{PVGW2017} are collections of real-world factual knowledge represented as RDF triples. A set of such triples typically constitutes a multi-relational graph with entities as nodes and relations as edges. The entities and relations often have rich textual information as their names and descriptions.
In recent years, KGs have been valuable resources in a variety of knowledge-intensive applications, such as question answering, search engines, and recommender systems.
Despite the increasing use, KGs often suffer from incompleteness with a high ratio of plausible facts missing \citep{farber2018linked}.
Knowledge Graph Completion (KGC) is then proposed to find these missing facts using the existing facts and/or external resources.

With the graph structure implied by triples, a large part of KGC methods uses KG embedding (KGE) techniques to encode the KG entities and relations into a vector space with their semantics like neighborhood graph patterns concerned, so that the missing facts can be inferred by their vector representations (a.k.a. embeddings) \citep{wang2017knowledge,chen2020knowledge}. We call them \textit{structure}-based methods (see Section \ref{sec:related_work} for more details).
Besides the graph structure, there are also some methods exploiting the text of the entities and relations as additional information for prediction, but methods of this type proposed before 2021 such as DKRL \citep{xie2016representation} (see their review \citep{gesese2021survey}) use non-contextualized text embeddings such as Word2Vec to encode the text, which specify a token a unique embedding and cannot fully capture its meaning in the context.

Recently, Pre-trained Language Models (PLMs) such as BERT \citep{devlin2019bert}, which encode the text with the tokens' contexts considered, have achieved great success in natural language processing (NLP), and they have been applied to KGC.
The methods often represent entities and relations using their textual information, view KGC as an NLP downstream task, and fine-tune PLMs to infer the missing facts.
For example, KG-BERT \citep{kgbert} takes as input a triple's whole text, encodes the text with BERT, and feeds the text encoding into a classifier to predict its plausibility (score);
KGT5 \citep{kgt5} leverages the Seq2Seq PLM framework to directly generate a triple's missing part conditioned on its other two known parts.
We call the methods that utilize the textual information with some PLMs as \textit{PLM}-based.
The challenge of these methods thus lies in incorporating the graph structure simultaneously.
Several efforts have been made in this direction.
For example, StAR \citep{star} and LASS \citep{lass} further forward the PLM-based text encoding to a KGE model to fine-tune the PLM and learn the structure embeddings jointly.
However, fine-tuning the PLM is \textit{(i)} costly in both computation and storage, limiting the applicability to larger models, and \textit{(ii)} more prone to overfit and forget the knowledge e.g. linguistic inherence learned during pre-training, in many cases limiting the performance \citep{gpt3,ding2023parameter}.


\begin{figure}[!t]
  \centering  \includegraphics[width=0.8\linewidth]{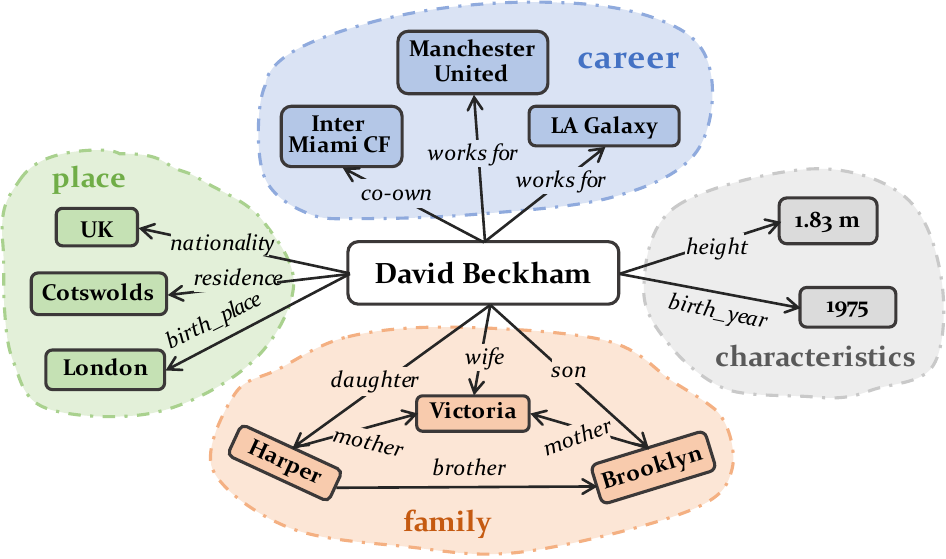}
\caption{An example of the KG entity ``David Beckham'' which is associated with neighboring entities of different aspects (e.g., ``family'', ``career'') by different kinds of relations. }
\label{fig:introduction}
\end{figure}

Meanwhile, a more efficient technology of using PLMs named \textbf{prompt-tuning} has arisen: adapting the PLM directly as a downstream task predictor by adding (learnable) task-specific prompts on the input side without fine-tuning the PLM \citep{gpt3,lester2021power,liu2023gpt}.
For example, the prompt ``It was $\texttt{[MASK]}$'' prepended to the input sentence ``No reason to watch it.'' formulates a binary sentiment classification task as the masked token prediction which is one of the pre-training tasks of LMs e.g. BERT. In general, such prompts on the one hand reduce the task gap between language modeling and downstream tasks with more generalization gains from PLMs, and on the other hand allow to freeze the PLMs with fewer parameters to tune.

Although prompt-tuning has been applied to various fields including NLP \citep{liu2023pre}, computer vision \citep{zhou2022learning}, protein design \citep{nathansen2023evaluating}, etc, relatively few attempts have been proposed for KGC.
The only one we know is CSProm-KG \citep{csprom_kg}, which uses a frozen PLM to incorporate the textual information into a KG's structural embeddings through structure-aware soft prompts, and then feeds the enhanced structural embeddings into a KGE model to predict the triple. However, the proposed structure-aware soft prompt is still quite preliminary and the whole method neglects the potential of the PLM for triple prediction.

In exploring prompt-tuning for KGC, we have the following questions: (1) \textit{How to make KGC close to the pre-training tasks of a PLM to obtain more task-relevant knowledge especially when the PLM is frozen?} (2) Given massive triples in a KG, \textit{how to effectively fuse the graph structure and the text by prompts?}
As shown in Fig. \ref{fig:introduction}, an entity in a KG is often connected to different neighboring entities via different relations, indicating semantics of different aspects. 
To complete the triple (\textit{David Beckham, member of team, ?}), the text about \textit{David Beckham}'s career should have higher attention to neighbors connected to \textit{David Beckham} through career-related relations such as \textit{works for} than to neighbors connected through other kinds of relations.
Thus to answer the second question, we need to learn the relevance between different text parts and different semantic aspects of the graph.
The self-attention mechanism of the PLM can automatically learn the relevance between two sequences. A straightforward idea thus is serializing the neighborhood into a sequence and feeding it to a self-attention layer together with the text \revise{\citep{chepurova2023better}}. However, it is impracticable due to PLMs' maximum sequence length limit.

To this end, we propose a new PLM-based KGC method \textbf{PDKGC} which includes a \textbf{hard task prompt} and a \textbf{disentangled structure prompt} for tackling the above two questions.
The hard task prompt is a pre-defined template containing the $\texttt{[MASK]}$ token, which reformulates the KGC task as a token prediction task in line with the pre-training tasks of many PLMs.
The disentangled structure prompt is a series of trainable vectors (i.e., soft prompts) generated from disentangled entity embeddings which are learned by a graph learner with selective aggregations over different neighboring entities.
It is prepended to the hard task prompt to form the inputs of a frozen PLM.
In this way, we incorporate structure knowledge into a PLM with shorter prompts and utilize powerful self-attention to learn the relevant part for a specific triple to complete.

After encoding by the PLM, for a KG triple to complete, PDKGC not only includes a \textit{textual predictor} to output a probability distribution over all the entities based on the encoded representation of the $\texttt{[MASK]}$ token which has fused relevant structural knowledge, but also includes a \textit{structural predictor} to simultaneously produce the entity probabilities by forwarding the encoded representations of structural soft prompts to a KGE model.
Naturally, their outputs can be further combined for a more comprehensive prediction.
Evaluations on \revise{three} popular KG datasets demonstrate the superiority of our proposed model to CSProm-KG, as well as fine-tuned \textit{PLM}-based methods and traditional \textit{structure}-based methods.

\section{Related Work}\label{sec:related_work}
\subsection{Structure-based Methods}\label{sec:references_struc}
These methods generally consist of three steps:
\textit{(i)} assigning a trainable embedding to each entity and relation, 
\textit{(ii)} defining a scoring function to measure the plausibility of a triple,
and \textit{(iii)} optimizing the entity and relation embeddings such that the positive triples get high scores while the negative ones get low scores.
According to how a triple is scored, existing methods can be grouped into three types: \textit{translation}-based ones such as TransE \citep{transe} and RotatE \citep{rotate}, \textit{semantic matching}-based ones such as DistMult \citep{distmult} and ComplEx \citep{complex}, and \textit{neural network}-based ones such as ConvE \citep{ConvE}.
Recently, graph neural networks (GNNs) are also employed to encode the entity by aggregation.
Typical practices include relation-aware GCN \citep{RGCN}, CompGCN \citep{CompGCN}, etc.
All of these models have shown their capability to embed complex semantics of the graph structure of a KG and have achieved promising results for KGC.

\subsection{PLM-based Methods}\label{sec:text_based_methods}
\textit{PLM}-based approaches use plain text (sequence of tokens) to represent the entities and relations for predicting the missing triples.
Some of them take PLMs as encoders, encoding the textual information and then predicting the plausibility of triples by feeding the encoded representations into a prediction layer.
More specifically, KG-BERT \citep{kgbert}, MTL-KGC \citep{mtl_kgc} and PKGC \citep{pkgc} pack the text of the head, relation and tail of a triple as one sentence, forward it into BERT, and feed the output at the $\texttt{[CLS]}$ token into a simple MLP layer to predict whether the triple is true or not.
BERTSubs \citep{chen2023contextual} is quite similar but is to predict the subsumption relation between two concepts in an ontology.
To predict the missing entity in an incomplete triple, these methods have to traverse all the entities to generate a set of candidate triples and predict all their scores.
To avoid this combinatorial explosion, StAR \citep{star} and SimKGC \citep{wang2022simkgc} separately encode the text of $(h, r)$ and $t$ using Siamese BERTs and predict the triple score by measuring the compatibility of these two encoded parts.
\revise{Following the same model architecture, SKG-KGC \citep{shan2024SKG-KGC} further \textit{i)} introduces the relation classification to perform multi-task learning together with the original entity prediction, \textit{ii)} extends the training set by packing the triples with identical $(h,r)$ or $(r,t)$ into one new triple.}
Nevertheless, they rely on \revise{massive and} high-quality negative samples.
Motivated by the masked token prediction in PLM pre-training, MEM-KGC \citep{mem_kgc} regards the missing entity of a triple as the masked token, and classifies this token over all the entities with the text of the known entity and relation.
LP-BERT \citep{lp_bert} proposes a two-stage model, which first predicts the masked entities, relations and partial tokens for pre-training and then contrastively matches the separately encoded $(h, r)$ and $t$ for fine-tuning.

Some methods take encoder-decoder or decoder-only PLMs to directly generate text of the missing entity to complete a triple.
KGT5 \citep{kgt5} makes the first attempt by pre-training a T5 model using large-scale KG datasets from scratch.
GenKGC \citep{genkgc} and KG-S2S \citep{kg_s2s} instead fine-tune BART and T5, respectively, with effective decoding strategies proposed.
Although these methods improve the inference efficiency by avoiding the traversal of all the candidates, the auto-regressive generation still takes a long time. Also, matching the generated text with the existing entities is non-trivial but challenging since an entity may have diverse surface names and may not exist in the KG.

To sum up, due to the inherent task gap between KGC and PLM pre-training, the above PLM-based methods require diverse fine-tuning strategies with both positive and negative samples, and mostly cost much time in training and inference.
Meanwhile, they mainly rely on the textual information alone, with the graph structure weakly incorporated.

\subsection{Joint Methods}

Before PLMs, there were already some KGC methods trying to utilize both structure and text knowledge for KGC \citep{xie2016representation,xu2017knowledge,kristiadi2019incorporating,gesese2021survey}.
However, they often adopt non-contextualized text embedding methods.
Thanks to the advances in PLMs, KEPLER \citep{wang2021kepler} proposes to use PLMs to encode entity descriptions as entity embeddings and then jointly optimize the KGE and masked language modeling objectives on the same PLM.
StAR \citep{star} additionally composes the text encodings of $(h,r)$ and $t$ using translation-based KGE methods' score functions, while LASS \citep{lass} follows KG-BERT to encode the full text of a triple but forwards the pooled text encodings of $h$, $r$, $t$ into a KGE model to reconstruct the KG structure.
\revise{Given $(h,r,?)$, \citep{chepurova2023better} follows KGT5 to predict the missing entity through sequence generation but extracts entities and relations adjacent to $h$ from the KG and verbalizes them as additional input. Furthermore, to avoid too long input sequences, the authors also sort the neighbors based on relation semantic similarity. In contrast to this naive idea, our PDKGC proposes to encode the neighborhood into disentangled structural embeddings with comprehensive but shorter and semantic-independent input introduced. Notably, the above methods all require to fine-tune the PLMs.}

Recently, \citep{csprom_kg} proposed \textbf{CSProm-KG}, which is the first work to investigate frozen PLMs for KGC.
It utilizes the text encoding from a frozen PLM for enhancing the structure embeddings by soft prompts, but predicts the missing triple using the structure embeddings alone by feeding them to a KGE model such as ConvE.
Our \textbf{PDKGC} also includes such a \underline{text-augmented KGE module} (i.e., the \textit{structure predictor}) but goes beyond it. It simultaneously has a \underline{structure-augmented text predictor} to enable the frozen PLM to predict an incomplete triple through a hard task prompt. 
These two predictors complement each other and provide more comprehensive predictions, leading to better results with a simple ensemble method.
Most importantly, CSProm-KG focuses on non-disentangled structural representations with limited attention between the text encoding and the graph structure embedding, while our PDKGC learns disentangled entity representations, through which more fine-grained correlations between the text and the graph structure can be learned for more robust prediction.

\begin{figure*}[!t]
  \centering
  \includegraphics[width=0.9\linewidth]{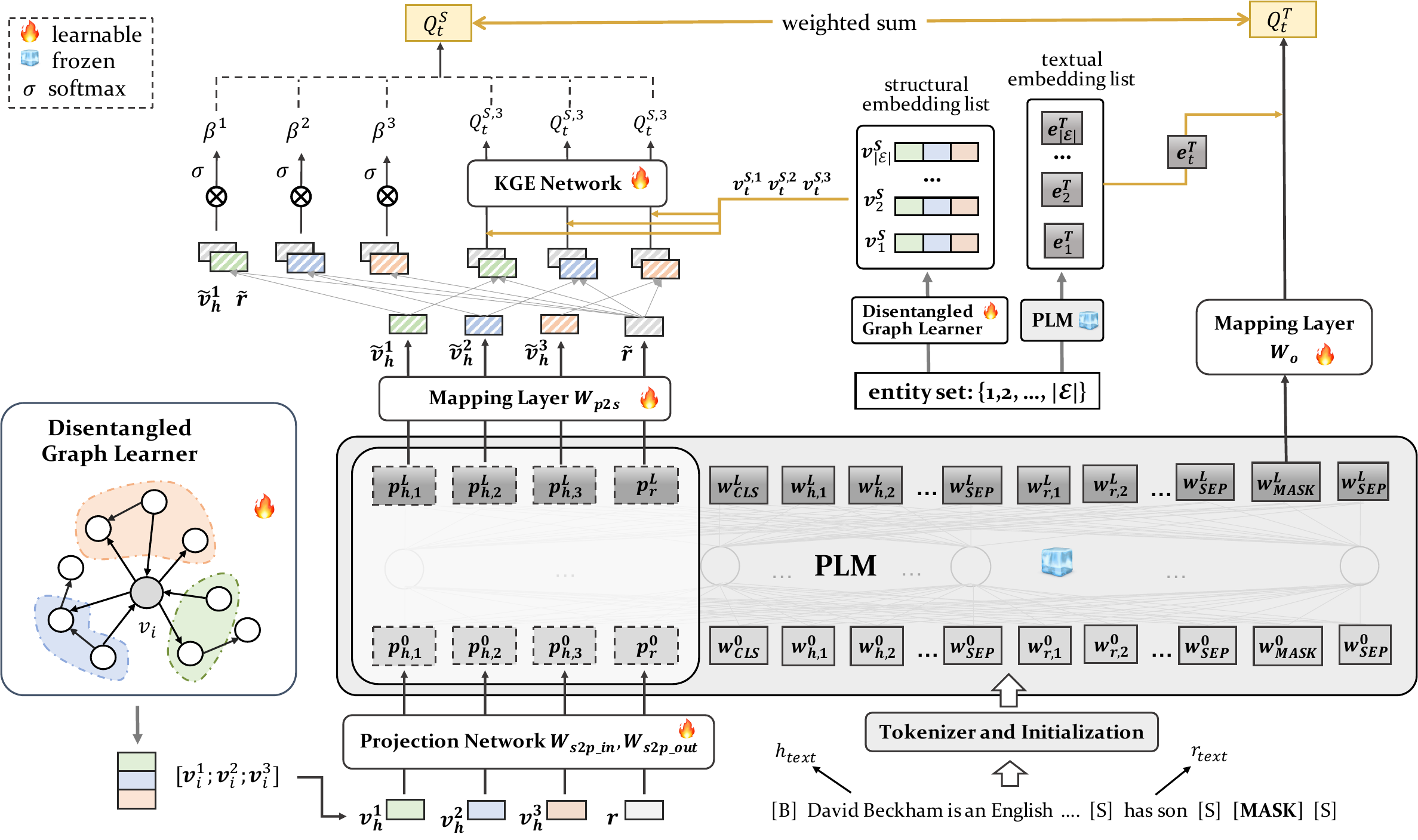}
\caption{The framework overview of our proposed PDKGC, with the KG triple example (\textit{David Beckham, has son, ?}) to complete. It includes (1) a disentangled graph learner that learns structure semantics of different $K$ aspects for each entity, here we take $K=3$ as an example, (2) a structure-aware text encoder that encodes the triple text together with a set of prefix prompts generated from disentangled structural embeddings, i.e., our proposed disentangled structure prompt with $\bm{p}^0$ and $\bm{p}^L$ at the first and last layers of the frozen PLM, respectively, and (3) two predictors that respectively take as input the structure-augmented textual encoding, i.e., $\texttt{[MASK]}$ token's hidden vector at PLMs' final layer $\bm{w}_{MASK}^L$, and the text-augmented structural encoding, i.e., the structural prompts at PLM's last layer $\bm{p}_{h,1}^{L}, \bm{p}_{h,2}^{L}, \bm{p}_{h,3}^{L}$ and $\bm{p}_{r}^L$, for predicting the probabilities (scores) that $t$ is the correct tail entity, i.e., $Q^S_t$ and $Q^T_t$, which can be further fused to output a final score. Notably, $\bm{w}_{h,i}^0$ is the input embedding of the $i$-th token in the head entity $h$'s textual names and descriptions, while $\bm{p}_{h,k}^{0}$ represents the input token sequence embedding corresponding to $h$'s $k$-th disentangled embedding.}
\label{fig:framework}
\end{figure*}

\section{Methodology}
In this section, we begin by first introducing the preliminary, including a formal definition of the KGC problem we aim at and an overview of the pre-trained language models (PLMs).
Then, as shown in Fig. \ref{fig:framework}, we introduce the hard task prompt applied in this study for reformulating KGC, and a disentangled graph learner for learning disentangled structural embeddings.
Based on them, we present a structure-aware text encoder built upon frozen PLMs for generating the disentangled structure prompt and encoding it with the text included in the hard task prompt.
Finally, two predictors based on textual and structural encodings are designed to simultaneously output the entity ranking results.

\subsection{Preliminary}
\vspace{0.15cm}
\subsubsection{KGC Problem Formulation}
In this study, a KG is formulated as $\mathcal{G} =\{ \mathcal{E}, \mathcal{R}, \mathcal{T}\}$, where $\mathcal{E}$ is a set of entities, $\mathcal{R}$ is a set of relations, and $\mathcal{T} = \{(h, r, t)| h, t \in \mathcal{E}; r \in \mathcal{R}\}$ is a set of relational facts in form of RDF\footnote{Resource Description Framework. See \url{https://www.w3.org/RDF/}.} triple.
For a triple $(h,r,t)$, $h$, $r$ and $t$ are called head entity, relation and tail entity, respectively.
For each entity or relation, it is often associated with a phrase of surface names and/or a paragraph of textual descriptions as its textual information. 
The completion is then defined to predict an input candidate triple as true or not, i.e., \textit{triple classification}, or predict the missing entity/relation in a triple with the other two elements given, i.e., \textit{entity/relation prediction} or \textit{link prediction}.
In our paper, we aim at training a model for the more challenging \textit{entity prediction} task for KGC, i.e., given a head $h$ (resp. tail $t$) and a relation $r$, the model is expected to find out a tail $t$ (resp. head $h$) from $\mathcal{E}$ for a new and correct triple $(h,r,t)$.
During inference, for an incomplete triple such as $(h, r, ?)$, the trained model will rank all the potential tail entities according to the probabilities of them being the correct or the scores of all candidate triples $\{(h, r,t')|t' \in \mathcal{E}, (h, r, t') \notin \mathcal{T}\}$. A tail entity is ranked at a higher position if it makes the current triple more plausible.

\subsubsection{Pre-trained Language Models}
PLMs are language models that are pre-trained on large-scale corpora in a self-supervised fashion, most of them derive from the Transformer \citep{vaswani2017attention} design, containing the encoder and decoder modules empowered by the self-attention mechanism.
Based on model architectures, PLMs can be grouped into encoder-only, encoder-decoder, and decoder-only.
Encoder-only PLMs only use the encoder to encode the input sequence, and are often pre-trained to predict the randomly masked tokens for recovery,
i.e., masked language modeling (MLM). Typical models include BERT \citep{devlin2019bert}, RoBERTa \citep{liu2019roberta} and ELECTRA \citep{clark2020electra}. After pre-training, an extra prediction layer, such as MLP, is often added to fine-tune the pre-trained models to solve downstream tasks.
Encoder-decoder PLMs include an encoder that encodes the input sequence into a hidden space, and a decoder that generates the target output text conditioned on these hidden states. Models in this group e.g. T5 \citep{raffel2020exploring}, BART \citep{lewis2020bart} and GLM \citep{zeng2023glm} are suitable for downstream tasks that generate text.
Decoder-only PLMs only employ the decoder to generate target text, and the pre-training paradigm is to predict the next word in a sentence.
Many state-of-the-art PLMs e.g. GPT-4 \citep{gpt4} and LLaMA \citep{touvron2023llama} follow this design.

\subsection{Hard Task Prompt}
As introduced above, many PLMs are pre-trained by the missing token prediction task, where a standard cross-entropy loss is applied to score the true token against all other tokens.
Therefore, to meet the needs of bridging KGC to these pre-training tasks, so as to distill the pre-trained knowledge from PLMs without having to fine-tune, we propose an ``auto-completion'' task prompt, to view the missing entity as a missing token and coax the off-the-shelf PLM models into producing a textual output based on the entity vocabulary. Formally, in the case of \textit{tail entity prediction}, the task prompt is defined as:
\begin{equation}
    X_{prompt} = \texttt{[B]} h_{text} \texttt{[S]} r_{text} \texttt{[S]} \texttt{[MASK]} \texttt{[S]}
\end{equation}
where $\texttt{[MASK]}$ is a placeholder to represent the missing tail entity $t$ to predict, the texts of the given head entity and relation, i.e., $h_{text}$ and $r_{text}$, are kept as the context.
$\texttt{[B]}$ and $\texttt{[S]}$ are special tokens used in PLMs, for encoder-only PLMs such as BERT, they are $\texttt{[CLS]}$ and $\texttt{[SEP]}$, for decoder-only PLMs such as LLaMA, $\texttt{[B]}$ is initialized as $\texttt{[BOS]}$, and the last $\texttt{[S]}$ is $\texttt{[EOS]}$.
$X_{prompt}$ is then fed into an arbitrary PLM, and the PLM will decide which entity is more appropriate to fill in $\texttt{[MASK]}$ by 
feeding its output hidden vector $\bm{w}_{\texttt{[MASK]}}^L$ into a liner layer parameterized by weights $\bm{W}_e \in \mathbb{R}^{|\mathcal{E}| \times H}$:
\begin{align}\label{eq:mask_prediction_ori}
\begin{split}
    p(t|X_{prompt}) &= p(\texttt{[MASK]}=t|X_{prompt}) 
     \\
     &= \frac{exp(\bm{e}_t \cdot \bm{w}_{\texttt{[MASK]}}^L)}{\sum_{t' \in \mathcal{E}} exp(\bm{e}_{t'} \cdot \bm{w}_{\texttt{[MASK]}}^L)}
\end{split}
\end{align}
where $H$ is the PLM's hidden vector size, $L$ is its layer number in total, and $\bm{e}_t$ is the row vector in $\bm{W}_e$ that corresponds to the correct entity $t$. Similarly, the cross-entropy loss is used to optimize the prediction, but instead scores the true entity against all other entities.
Notably, for encoder-decoder and decoder-only PLMs, we only decode one token corresponding to the predicted entity without having to auto-regressively generate a text sequence for representing $t$, which would cost extra inference time.
The case of \textit{head entity prediction} is the same except that the missing head entity is replaced by $\texttt{[MASK]}$ for prediction. 
Without losing the generalization, in the remainder of this section, we use the case of \textit{tail entity prediction} to introduce the method.
Next, we will introduce how to leverage the disentangled structure prompt to incorporate highly relevant local structural information into the PLM.

\subsection{Disentangled Graph Learner}
Given the graph context of a triple to complete, i.e., $(h, r, ?)$, especially the surrounding triples of $h$, we find that only a subset of neighbors carries valuable information that can be used to augment the inference, i.e., highly relevant graph structures.
Therefore, we propose to learn a disentangled representation (embedding) for each entity. 
Such a representation contains multiple components, each of which encodes the features of a specific subset of neighbors that are highly relevant to this entity in a certain semantic aspect. 
Briefly, different components correspond to different semantic aspects of the entity.
Take the entity \textit{David Beckham} as an example, its associated relations \textit{daughter} and \textit{wife} and its neighboring entities connected by these two relations represent semantics of the family aspect, and they are expected to be encoded in one component of this entity's disentangled representation.
In contrast, the semantics of the career aspect of \textit{David Beckham} are expected to be encoded in another component.
With such a disentangled representation, more relevant semantics from the graph can be captured for the contextual text of the triple to complete.
To better understand, in this subsection, we use $v_i$ to denote an entity with its index $i$, and its disentangled representation is denoted as $\bm{v}_i = [\bm{v}^1_i, \bm{v}^2_i, ..., \bm{v}^K_i]$, where $\bm{v}^k_i \in \mathbb{R}^d$ denotes the $k$-th component, $d$ is its embedding size, and $K$ is the number of components.

To identify the aspect-specific subset, we follow the attention-based neighborhood routing strategy proposed in disentangled graph learning \citep{ma2019disentangled,wu2021disenkgat}.
Also, considering the various relation types in the neighborhood, we apply a \textbf{relation-aware attention mechanism}.
Specifically, for the $k$-th aspect, the attention value of one neighbor $v_j$ of entity $v_i$ is computed by the similarity of the $k$-th component embeddings of $v_j$ and $v_i$ in the subspace of their relation $r$ following the assumption that when a neighbor contributes more to $v_i$ during encoding, their relation-aware representations are more similar, formally:
\begin{align}\label{eq:attention}
\alpha_{(v_i,r,v_j)}^{k} &= softmax((\bm{v}_{i,r}^k)^T \cdot \bm{v}_{j,r}^k)
    \nonumber \\
&= \frac{exp((\bm{v}_{i,r}^k)^T \cdot \bm{v}_{j,r}^k)}{\sum_{(v_{j'}, r')\in \mathcal{N}(v_i)} exp((\bm{v}_{i,r'}^k)^T \cdot \bm{v}_{j',r'}^k)} \\
\bm{v}_{i,r}^k &= \bm{v}_{i}^k \circ \bm{W}_r, \;\;\; \bm{v}_{j,r}^k = \bm{v}_{j}^k \circ \bm{W}_r 
\end{align}  
where 
$\bm{v}_{i,r}^k$ is the $k$-th component embedding of $v_i$ w.r.t. relation $r$, $\circ$ denotes the Hadamard product, and $\bm{W}_r$ is a learnable projection matrix of $r$ for projecting component embeddings into a relation specific subspace.
$\mathcal{N}(v_i)$ denotes neighboring entity-relation pairs, for each of which the entity is associated with $v_i$ through a specific relation, including $v_i$ itself with a special self-connection relation. The dot-product similarity is adopted here.

With attention values, we next aggregate the highlighted neighbors' features to learn each component so as to encode the relevant graph structures into component embeddings:
\begin{equation}
\bm{v}_{i}^{k,l} = \sigma(\sum_{(v_j,r)\in \mathcal{N}(v_i)} \alpha_{(v_i,r,v_j)}^{k,l-1} \phi (\bm{v}_{j}^{k,l-1}, \bm{r}^{l-1}, \bm{W}_r))
\end{equation}
here we add $l$ in the superscript of $\bm{v}_{i}^{k}$ to denote the hidden state of $v_i$'s $k$-th component after $l$ layers aggregation, and $l\in \{1, ..., L_{g}\}$ with $L_{g}$ as the total number of aggregation layers.
$\bm{r}^{l-1}$ is relation $r$'s embedding in the $(l-1)$-th layer, and we also update it with layer-specific transformation matrix parameterized by $\Theta_r^{l-1}$ as: $\bm{r}^{l} = \bm{r}^{l-1}\cdot \Theta_r^{l-1}$.
$\phi$ is a combination operator for fusing the features of neighboring entities and relations.
Here, we refer to \citep{wu2021disenkgat} to implement it via e.g. crossover interaction.
$\bm{v}_{i}^{k,0}$ and $\bm{r}^{0}$ are randomly initialized, and
$\bm{v}_{i}^{k,L_{g}}$ is outputted at the last layer which has encoded the neighborhood information specific to $k$-th aspect. In the remainder, we use $\bm{v}_i^k$ to denote $\bm{v}_{i}^{k, L_{g}}$ and $\bm{r}$ to denote $\bm{r}^{L_{g}}$ for simplicity.

\subsection{Structure-aware Text Encoder}\label{sec:structure_aware_text_encoder}
After obtaining the embedded structural information, we next integrate them with the PLM.
Specifically,
we learn a projection network to translate the learned entity and relation embeddings into a sequence of token embeddings, and prepend them at the inputs to the frozen PLMs.
Before that, given the facts that not all the structure knowledge is truly useful for a target triple, and we have learned disentangled representations for entities with their different semantic aspects considered, a question raised here is: how to select the entity embedding components that are highly related to the triple to complete.
A practicable solution is to measure the semantic relatedness between a component and the relation in the target triple by e.g. computing the similarity of their structural embeddings, and select those highly related.
However, the structural representations might not be comprehensive enough to accurately evaluate the relatedness.
Thus, we propose to feed all the components into the PLM and rely on the powerful self-attention mechanism to make decisions, where the textual information of the current triple also helps in determining which components are more relevant.

Specifically, for entity $v_i$ and its one component embedding $\bm{v}_i^k$, we first generate a set of token vectors for $\bm{v}_i^k$ as:
\begin{equation}
    \bm{p}_{i,k} = \bm{W}_{s2p\_out} \cdot (ReLU(\bm{W}_{s2p\_in} \cdot \bm{v}_i^k))
\end{equation}
where $\bm{W}_{s2p\_in} \in \mathbb{R}^{d_h \times d}$ and $\bm{W}_{s2p\_out} \in \mathbb{R}^{(H \ast n) \times d_h}$ are weight matrices in the two-layer projection network, $d_h$ is its middle hidden size, $d$ denotes the embedding size of each component.
The projected output is then re-shaped as $\mathbb{R}^{H \times n}$, where $H$ represents the hidden state size in the PLM and $n$ is the length of the generated prompts, meaning that we translate each component into a sequence of length $n$.
The relation embeddings are also translated similarly.
Consequently, for a target triple $(h,r,?)$ with the tail entity missing, we generate for the corresponding structural embeddings $\bm{v}_h^1, \bm{v}_h^2, ..., \bm{v}_h^K$ and $\bm{r}$ a sequence $\bm{p}=[\bm{p}_{h,1}; \bm{p}_{h,2}; ...; \bm{p}_{h,K}; \bm{p}_r]$, with $(K+1)\ast n$ tokens in total, where $[;]$ denotes the vector concatenation operation.
$\bm{p}$ is thus the \textbf{disentangled structure prompt}.
Meanwhile, we also convert the $h_{text}$ and $r_{text}$ in the hard task prompt into corresponding input embeddings with the PLM's tokenizer and pre-trained token embedding table.
As shown in Fig. \ref{fig:framework}, the first token of $h_{text}$ is represented as $\bm{w}_{h,1}$, with $\bm{w}_{h,1}^0$ and $\bm{w}_{h,1}^L$ as the input of the first layer and the output of the last layer, respectively, of the PLM.
The disentangled structure prompt sequence is then re-denoted as $\bm{p}^0 = [\bm{p}_{h,1}^{0}; \bm{p}_{h,1}^{2}; ...; \bm{p}_{h,K}^{0}; \bm{p}_r^0]$ and prepended to the input embeddings to forward to the PLM.

In practice, to avoid too long inputs for PLMs, especially when we load all components of an entity, we follow \citep{li2021prefix,csprom_kg} to build layer-wise prompts with shorter lengths but inserted at each layer.
To be more specific, we modify the output matrix of the projection network as $\bm{W}_{s2p\_out} \in \mathbb{R}^{(L \ast H \ast n) \times d_h}$, where $L$ means the number of layers in a PLM.
In this way, the prompt $\bm{p}$ with relatively small $n$ is prepended at the beginning of each layer and frequently interacts with the textual information.

To sum up, we implement the structure-aware text encoding by prepending the disentangled structure prompts to the text encoder, which introduces more relevant structure knowledge in the missing token's final representation (i.e., $\bm{w}_{\texttt{[MASK]}}^L$).
On the other hand, we also leverage the semantics implied in text to help the disentangled graph learning through self-attention over all components of the entity representation, and get a set of text-enhanced structure representations (i.e., $\bm{p}_{h,k}^{L}$ and $\bm{p}_r^L$).

\subsection{Textual and Structural Predictors}
With the final hidden vector of the $\texttt{[MASK]}$ token $\bm{w}_{\texttt{[MASK]}}^L$, we next predict the missing entities.
In our preliminary experiments, we follow Eq. \eqref{eq:mask_prediction_ori} to use a randomly initialized linear layer parameterized by $\bm{W}_e$ to classify $\bm{w}_{\texttt{[MASK]}}^L$ over all entities, where the $i$-th row vector $\bm{e}_i$ serves as the classification vector corresponding to the $i$-th entity.
Feeding $\bm{w}_{\texttt{[MASK]}}^L$ into the linear layer can thus be viewed as matching it with these classification vectors, the most similar one whose corresponding entity is the predicted entity. 
However, these classification vectors are directly learned representations instead of encoding the textual information like $\bm{w}_{\texttt{[MASK]}}^L$.
Therefore, to better utilize the textual semantics, for each candidate entity, we run a frozen textual encoder, which is the same as the main PLM, to encode its textual information and obtain a fixed textual embedding $\bm{e}_i^T$ in advance. Then, we replace $\bm{W}_e$ with these textual embeddings and predict the probability of the $i$-th entity being the correct entity as: 
\begin{equation}\label{eq:mask_prediction}
    Q^T_i = \bm{e}_i^T \cdot \bm{W}_o \cdot \bm{w}_{\texttt{[MASK]}}^L
\end{equation}
where $\bm{W}_o \in \mathbb{R}^{H \times H}$ is a trainable transformation matrix for more flexible learning.

With \textit{tail entity prediction}, the standard cross-entropy loss is applied to train the whole model as:
\begin{equation}
\begin{aligned}
\mathcal{L}^{T} = -\frac{1}{B}\sum_{(h, r) \in batch} ((1-\epsilon) \cdot \text{log} \ p(t|\widetilde{X}_{prompt}) + \\ \frac{\epsilon}{|\mathcal{E}|-1} \sum_{t' \in \mathcal{E}/\{t\}} \text{log} \  p(t'|\widetilde{X}_{prompt}))
\end{aligned}
\end{equation}
where $\widetilde{X}_{prompt}$ extends $X_{prompt}$ with the disentangled structural prompts. 
$p(t|\widetilde{X}_{prompt}) = \frac{exp(Q_t^T)}{\sum_{t' \in \mathcal{E}} exp(Q_{t'}^T)}$, and $t$ is the label (ground-truth tail entity) of the given testing triple $(h,r,?)$. $B$ is the batch size, $\epsilon$ is for label smoothing often set to $0.1$.

We call the prediction made by Eq. \eqref{eq:mask_prediction} \textit{textual predictor} since it mainly relies on the LMs for inference.
Furthermore, to ensure that the structural embeddings really capture the dependencies on graph, we propose a parallel \textit{structural predictor} that forwards the structural embeddings especially the text-enhanced structural embeddings to a KGE model to reconstruct the graph structures.
Specifically, we extract the hidden vectors of structural prompts at the last layer and map them into the graph embedding space again through a linear layer parameterized by $\bm{W}_{p2s} \in \mathbb{R}^{d \times (H \ast n)}$.
Then, regarding that we have multiple mapped entity prompts with different semantic aspects, and the importance of each component has been weighted by the current triple, we conduct component-level prediction and leverage a triple-wise attention mechanism to fuse the results from different components.

Take the \textit{tail entity prediction} $(h,r,?)$ as an example, once obtained the mapped head entity component embedding $\bm{\widetilde{v}}_h^k$ and the mapped relation embedding $\bm{\widetilde{r}}$, for an entity $v_i \in \mathcal{E}$, we can choose a triple score function from one of the existing geometric KGE models to compute its score to be the correct tail entity. Here we take the score function of TransE \citep{transe} as an illustration:
\begin{equation}
    Q^{S,k}_{i} = \gamma - || \bm{\widetilde{v}}_h^k + \bm{\widetilde{r}} - \bm{v}^{S,k}_{i} ||_2^2
\end{equation}
where $\gamma$ represents a margin parameter controlling the score difference between true and false entities, $\bm{v}^{S,k}_i$ is the $k$-th component of the embedding of $v_i$ learned by the Disentangled Graph Learner, here we add ``S'' in the superscript of $\bm{v}^{k}_i$ to make a distinction between it and $v_i$'s corresponding textual embedding $\bm{e}^{T}_i$.

Meanwhile, since the relation information plays an important role in distinguishing the semantic aspects of entities, we obtain the attention weight of each prediction by computing the similarity between the mapped entity component prompts and relation prompts, formally:
\begin{equation}\label{eq:xxx}
\begin{aligned}
\beta_{(h,r)}^{k} &= softmax((\bm{\widetilde{v}}_h^k)^T \cdot \bm{\widetilde{r}}) \\ 
&=\frac{exp((\bm{\widetilde{v}}_h^k)^T \cdot \bm{\widetilde{r}})}{\sum_{k' \in \{1, 2, ..., K\}}exp((\bm{\widetilde{v}}_h^{k'})^T \cdot \bm{\widetilde{r}})} 
\end{aligned}
\end{equation}  

The final score of $v_i$ is computed as:
\begin{equation}
    Q_i^S = \sum_{k \in \{1, 2, ..., K\}} \beta_{(h,r)}^{k} Q_{i}^{S,k} 
\end{equation}
, and the standard cross-entropy loss is calculated as:
\begin{equation}
\begin{aligned}
\mathcal{L}^{S} = -\frac{1}{B}\sum_{(h, r) \in batch} ((1-\epsilon) \cdot \text{log} \ p(t|h,r) + \\ \frac{\epsilon}{|\mathcal{E}|-1} \sum_{t' \in \mathcal{E}/\left\{t\right\}} \text{log} \  p(t'|h,r))
\end{aligned}
\end{equation}
where $p(t|h,r) = \frac{exp(Q_t^S)}{\sum_{t' \in \mathcal{E}} exp(Q_{t'}^S)}$.

As we can see, we predict the tail entity using the structural prompts outputted from the PLM, which are representation enhanced by the triple's textual information, instead of the structural embeddings before feeding into the PLM.
This is similar to the textual predictor in Eq. \eqref{eq:mask_prediction} where the final representation of the $\texttt{[MASK]}$ token has been augmented by the triple's structure information. 
In this way, we not only obtain a model that can effectively fuse the textual and structural knowledge, but also have two predictions for a triple to complete.
Consequently, we can fuse them during inference by e.g., performing a weighted sum of the predicted scores.

The final training loss is defined as:
\begin{equation}
    \mathcal{L} = f (\mathcal{L}^{T}, \mathcal{L}^{S}) + \lambda \cdot \mathcal{L}_{mi}
\end{equation}
where $\mathcal{L}_{mi}$ is a mutual information based loss used to regularize the independence among disentangled components, $\lambda$ is its corresponding hyperparameter.
The function $f$, derived from \citep{kendall2018multi} for effective multi-task learning, contains a two-valued trainable parameter to automatically weight and sum the loss items of the \textit{textual predictor} and \textit{structural predictor}. See more details in our published codes.
During inference, we also use the well-trained $f$ to compute the weighted sum of the scores of these two predictors. In the future, it is expected to explore more effective score fusion solutions.

\section{Experiments}

\subsection{Experiement Settings}
\vspace{0.15cm}
\subsubsection{Datasets and Evaluation Metrics}
We experiment with two benchmark datasets that are widely used in the KGC domain, i.e., WN18RR \citep{ConvE} extracted from WordNet \citep{miller1995wordnet} and FB15K-237 \citep{toutanova2015observed} extracted from Freebase \citep{bollacker2008freebase}.
WordNet is a large lexical KG of English, where nouns, verbs, adjectives and adverbs are organized into sets of synonyms, each representing a lexicalized entity. Semantic relations such as hypernym, hyponymy and meronymy are used to link these entities.
Freebase is a collaboratively created KG for structuring human knowledge, which has been widely used to support KG-related applications, such as open information extraction and open-domain question answering \citep{jiang2019freebaseqa}.
\revise{Besides, we also conduct evaluations on a recently proposed KGC dataset CoDEx \citep{safavi2020codex} extracted from Wikidata \citep{vrandevcic2014wikidata} and Wikipedia, which improves upon existing KGC benchmarks in scope and level of difficulty. We adopt the largest version CoDEx-L with more entities and relations.}
The statistics of these \revise{three} datasets are given in Table \ref{tab:datasets}.


For entity text information, we use synonym definitions for WN18RR, and names and descriptions from  \citep{xie2016representation} for FB15K-237, following KG-BERT.
\revise{For CoDEx-L, we use Wikidata labels and descriptions, plus first paragraph from the Wikipedia pages.}
For the relations of these \revise{three} datasets, we use relation names as the textual information.

We evaluate our method PDKGC and the baselines with both \textit{head entity prediction} and \textit{tail entity prediction}.
For a triple to complete, i.e., $(h,r,?)$ or $(?,r,t)$, the methods are expected to rank a set of candidate entities in descending order according to their predicted scores as the correct head or tail entity of this triple.
A smaller rank of the ground truth entity indicates a better method.
Accordingly, we report two widely used metrics \citep{wang2017knowledge}: Mean Reciprocal Ranking (MRR), i.e., the average reciprocal rank over all test triples, and Hits@\{1,3,10\}, i.e., the ratio of testing triples whose ground truth entities are ranked within top 1/3/10.
Notably, these two metrics are both counted under the \textit{filter setting} \citep{transe} where other correct entities are filtered before ranking and only the current test one is left.

\subsubsection{Baselines and Variants of PDKGC}

We compare our methods with the existing \textit{structure}-based and \textit{PLM}-based methods, as well as those jointly encoding structural and textual information through PLMs.

\textbf{\textit{Structure}-based} methods include well-known TransE \citep{transe}, DistMult \citep{distmult} and ConvE \citep{ConvE}, which all define effective score functions for learning meaningful structural entity and relation embeddings.
Besides, we also make comparisons with two GNN-based methods CompGCN \citep{CompGCN} and DisenKGAT \citep{wu2021disenkgat}, which aim to capture richer structural knowledge for entities from their local neighborhood.
Especially, DisenKGAT is a method that learns disentangled representations for entities but only utilizes the graph structure for KGC.

\textbf{\textit{PLM}-based} methods present various solutions for KGC problems by incorporating textual information and taking PLMs as encoders or generators.
For each kind of these methods, we select one or two representative ones for comparison, including KG-BERT \citep{kgbert} and MTL-KGC \citep{mtl_kgc} which encode the full text of a triple, StAR \citep{star} \revise{and SKG-KGC \citep{shan2024SKG-KGC}} built upon separated encodings of $(h, r)$ and $t$,  
KGT5 \citep{kgt5} and KG-S2S \citep{kg_s2s} which generate text of the missing entity in a triple token by token,
and MEM-KGC \citep{mem_kgc} which predicts the missing entity through masked entity prediction.
These KGC methods all fine-tune PLMs.

\textbf{\textit{Joint}} methods embed and utilize both structure and text, here we \revise{focus on comparing} with those based on PLMs, including KEPLER \citep{wang2021kepler}, StAR \citep{star}, LASS \citep{lass} CSProm-KG \citep{csprom_kg} \revise{and KGT5+Neighbors \citep{chepurova2023better}}.
In our experiments, we mainly consider StAR, CSProm-KG \revise{and KGT5+Neighbors}.
KEPLER is a work at an early stage. It does not experiment with any of our \revise{three} datasets, and is hard to re-implement since the model is per-trained using a large-scale KG extracted from Wikidata.
Moreover, it performs worse than KGT5 on the corresponding test data, and KGT5 is already included in the baselines.
We also omit LASS since the results of important metrics of MRR, Hits@1 and Hits@3 are missing in the original work, and a lot of computation ($8$ V100 GPUs used in the original work) is required to re-implement them.
For CSProm-KG, we also compare one of its variants where the local adversarial regularization (LAR) module is removed, denoted as $\text{CSProm-KG}_{\text{non-LAR}}$.
LAR is mainly for distinguishing textually similar entities.
Formally, given a target triple $(h, r, ?)$ and the ground truth entity $t$, it selects a set of entities that are textually similar to $t$ as negative samples and uses a margin loss to constrain the distance between them.
\revise{Since KGT5+Neighbors also does not experiment with any of our three datasets, we try to re-implement it following the configurations released in the original paper.}

\begin{table}[htbp]
\centering
\caption{Summary Statistics of the Datasets.
}\label{tab:datasets}
\begin{tabular}{l|ccccc}
\toprule[0.7pt]
\textbf{Dataset}    & \textbf{$|\mathcal{E}|$} & \textbf{$|\mathcal{R}|$} & \textbf{$|\mathcal{T}_{train}|$} & \textbf{$|\mathcal{T}_{valid}|$} & \textbf{$|\mathcal{T}_{test}|$} \\\midrule[0.5pt]
WN18RR    
&  40,943 & 11  &   86,835  & 3,034  & 3,134  \\\midrule[0.5pt]
FB15K-237  
& 14,541 & 237 & 272,115  &  17,535 & 20,466       
\\\midrule[0.5pt]
\revise{CoDEx-L}
& \revise{77,951} & \revise{69} & \revise{551,193} & \revise{30,622} & \revise{30,622}

\\
\bottomrule[0.7pt]
\end{tabular}
\end{table}

\begin{table*}[htbp]
\centering
\caption{\small Overall Results \revise{on WN18RR and FB15K-237}. The best results are in bold and the second best results are underlined. Among baselines, the numbers in italic mean the results implemented by us, others are derived from the original papers.
}
\label{tab:overall_results}
\resizebox{0.9\linewidth}{!}{
\begin{tabular}{c|l|cccc|cccc}
\toprule[0.7pt]
\multicolumn{1}{c|}{\multirow{2}{*}{\textbf{Category}}}
& 
\multicolumn{1}{c|}{\multirow{2}{*}{\textbf{Methods}}}
&
\multicolumn{4}{c|}{\textbf{WN18RR}} & \multicolumn{4}{c}{\textbf{FB15K-237}} 
\\
& &  MRR &  Hits@1 & Hits@3 & Hits@10 &  MRR  & Hits@1 & Hits@3 &  Hits@10 
\\
\midrule[0.5pt]
\multicolumn{1}{c|}{\multirow{5}{*}{Structure-based}}
& TransE
& 0.243 & 0.043 & 0.441 & 0.532
& 0.279 & 0.198 & 0.376	& 0.441
\\
& DistMult
& 0.444  & 0.412 & 0.470 & 0.504
& 0.281 & 0.199 & 0.301 & 0.446
\\
& ConvE 
  & 0.456  & 0.419 & 0.470 & 0.531 
 &0.312  & 0.225 & 0.341 & 0.497
\\
& CompGCN
& 0.479 & 0.443 & 0.494 & 0.546
& 0.355 & 0.264  & 0.390  & 0.535
\\
& DisenKGAT
& 0.486  & 0.441 & 0.502 & 0.578
& \textit{0.366} & \textit{0.271} & \textit{0.405} & \textit{0.553}
\\
\midrule[0.5pt]
\multicolumn{1}{c|}{\multirow{9}{*}{PLM-based}}
& 
KG-BERT
&0.216 &0.041 &0.302 &0.524
&- & -& -& 0.420
\\
& MTL-KGC
& 0.331 &0.203 &0.383 &0.597
& 0.267 &0.172 &0.298 &0.458
\\
& StAR 
&0.401 &0.243 &0.491 & \revise{0.709}
& 0.296 &0.205 &0.322 &0.482
\\
& MEM-KGC
& 0.533 &0.473&0.570&0.636 
& 0.339	& 0.249	&0.372	&0.522
\\
& $\text{MEM-KGC}_{\text{fine-tuned}}$
& \it{0.530} & \it{0.483} & \it{0.559} & \it{0.611}
& \it{0.332}& \it{0.244} & \it{0.367} & \it{0.508}
\\
& $\text{MEM-KGC}_{\text{frozen}}$
& \it{0.264} & \it{0.193} & \it{0.300} & \it{0.401}
&\it{0.245}&\it{0.179}&\it{0.265}&\it{0.376}
\\
& KGT5
&0.508 &0.487 &- &0.544
&0.276 &0.210 &- &0.414
\\
& KG-S2S
&0.574 & \revise{\underline{0.531}} & 0.595 &0.661
& 0.336 &0.257 &0.373 &0.498
\\
& \revise{SKG-KGC}
& \revise{\textbf{0.722}} & \revise{\textbf{0.670}} & \revise{\textbf{0.751}} & \revise{\textbf{0.816}}
& \revise{0.350} & \revise{0.264} & \revise{0.377} & \revise{0.522}
\\
\midrule[0.5pt]
\multicolumn{1}{c|}{\multirow{3}{*}{Joint}}
& CSProm-KG
& \revise{0.575} & \revise{0.522} & 0.596 &0.678
&0.358 &0.269 &0.393 &0.538
\\
& $\text{CSProm-KG}_{\text{non-LAR}}$
&0.534& 0.489& - &0.624
& \textit{0.350} & \textit{0.259} & \textit{0.384}& 
\textit{0.530}
\\
& \revise{KGT5+Neighbors}
& \revise{\it 0.202} & \revise{\it 0.151} & \revise{\it 0.235} & \revise{\it 0.304}
& \revise{\it 0.168} & \revise{\it 0.123} & \revise{\it 0.185} & \revise{\it 0.269}
\\
\midrule[0.5pt]
\multicolumn{1}{c|}{\multirow{7}{*}{Ours}}
& 
$\text{PDKGC}_{\text{BERT}}$[T]
& 0.531 & 0.450 & 0.571 & 0.688
& 0.348 & 0.260 & 0.383 & 0.526
\\
& $\text{PDKGC}_{\text{BERT}}$[S]
& 0.543 & 0.490 & 0.565 & 0.646
& 0.370 & 0.278 & 0.407 & 0.551
\\
& $\text{PDKGC}_{\text{BERT}}$[C] 
& 0.568 & 0.500 & \revise{0.598} & 0.702
& \textbf{0.381} & \textbf{0.289} & \textbf{0.418} & \textbf{0.567}
\\
\cmidrule[0.3pt]{2-10}
& $\text{PDKGC}_{\text{RoBERTa}}$[T]
& 0.541 & 0.455 & 0.586 & 0.708
& 0.353 & 0.264 & 0.385 & 0.531
\\
& $\text{PDKGC}_{\text{RoBERTa}}$[S]
& 0.551 & 0.492 & 0.579 & 0.664
& 0.365 & 0.275 & 0.399 & 0.545
\\
& $\text{PDKGC}_{\text{RoBERTa}}$[C]
& \revise{\underline{0.577}} &0.505&\revise{\underline{0.609}}& \revise{\underline{0.713}}
&\underline{0.379}& \underline{0.285}&\underline{0.415}& \underline{0.566}
\\
\bottomrule[0.7pt]
\end{tabular}
}
\end{table*}

\begin{table}[htbp]
\centering
\caption{\small \revise{Overall Results on CoDEx-L. Among baselines, the numbers in italic mean the results implemented by us, others are derived from CoDEx's original paper \citep{safavi2020codex}. }
}
\label{tab:Results_CoDEx}
\resizebox{0.9\linewidth}{!}{
\begin{tabular}{l|cccc}
\toprule[0.7pt]
\multicolumn{1}{c|}{\multirow{2}{*}{\textbf{Methods}}}
&
\multicolumn{4}{c}{\textbf{CoDEx-L}} 
\\
&  MRR &  Hits@1 & Hits@3 & Hits@10 
\\
\midrule[0.5pt]
\footnotesize{TransE}
&0.187 & 0.116 & 0.219 &0.317
\\
\footnotesize{ConvE}
& 0.303 & 0.240 & 0.330 & 0.420
\\
\footnotesize{TuckER}
& 0.309 & 0.244 & 0.340 & 0.430
\\
\footnotesize{DisenKGAT}
& \it{0.324} & \it{0.252} & \it{0.359} & \underline{\it 0.456}
\\\midrule[0.5pt]
\footnotesize{MEM-KGC}
& \it{0.275} & \it{0.214} & \it{0.302} & \it{0.391}
\\
\footnotesize{KG-S2S}
& \it{0.244} & \it{0.192} & \it{0.268} &\it{0.353}
\\\midrule[0.5pt]
\footnotesize{$\text{CSProm-KG}_{\text{non-LAR}}$}
& \it{0.315} & \it{0.263} & \it{0.342} & \it{0.412} 
\\\midrule[0.5pt]
\footnotesize{$\text{PDKGC}_{\text{BERT}}$[T] }
& \underline{0.331} & \underline{0.266} & \underline{0.362} & 0.452 
\\
\footnotesize{$\text{PDKGC}_{\text{BERT}}$[S] }
& 0.321 & 0.260 & 0.347 & 0.436 
\\
\footnotesize{$\text{PDKGC}_{\text{BERT}}$[C] }
& \textbf{0.348} & \textbf{0.281} & \textbf{0.379} & \textbf{0.475} 
\\
\bottomrule[0.7pt]
\end{tabular}
}
\end{table}

\textbf{Variants of our \textit{PDKGC}.}
Among \textit{PLM}-based and \textit{Joint} baselines, we find that many of them are built upon PLMs of BERT \citep{devlin2019bert} and RoBERTa \citep{liu2019roberta}, e.g., KG-BERT and MEM-KGC fine-tune BERT-base, StAR reports the results with BERT-base, RoBERTa-base and RoBERTa-large, while CSProm-KG focuses on BERT-large.
Therefore, for fairer comparisons, we use BERT-large and RoBERTa-large as the frozen PLMs in our PDKGC, leading to two variants named $\text{PDKGC}_{\text{BERT}}$ and $\text{PDKGC}_{\text{RoBERTa}}$.
Notably, we use their uncased English versions.
In this case, our proposed hard task prompt formulates KGC as a masked token prediction task, which is in line with the masked language modeling (MLM) proposed for pre-training BERT and RoBERTa.
Correspondingly, MEM-KGC is very close to us but fine-tunes BERT-base, therefore, we also re-produce it with both fine-tuned and frozen BERT-large, denoted as $\text{MEM-KGC}_{\text{fine-tuned}}$ and $\text{MEM-KGC}_{\text{frozen}}$, respectively.

Regarding our proposed \textit{textual predictor} and \textit{structural predictor} both yield valid entity ranking results, we also record them together with the ensemble ones.
We use the form of ``X[Y]'' to distinguish, where ``X'' means the model variant, ``Y=\{T, S, C\}'' represent the results from \textit{textual predictor}, \textit{structural predictor}, and the ensemble results, respectively.

\subsubsection{Implementation Details}
For more convenient training and inference, in our paper, we follow previous works e.g. \citep{ConvE,CompGCN} to add an inverse triple $(t, r^{-1}, h)$ for each triple $(h,r,t)$ to predict the head entity, where $r^{-1}$ is the inverse relation of $r$. Based on such reformulation, we only need to deal with the \textit{tail entity prediction} problem.
For inverse relation $r^{-1}$, we add a prefix word “reverse” to the text of $r$. For examples, if $r$ has the name of ``produced\_by'', then $r^{-1}$ is named as ``reverse: produced\_by''.
As for KGE algorithms, we by default use ConvE \citep{ConvE} and accordingly calculate the triple score as the dot-product between the representation of $(h, r)$ by 2D convolutions and the tail entity representation.
In Table \ref{tab:KGE_results}, we also have ablation study results with different KGE methods.

All the experiments are run on a single NVIDIA Tesla V100 GPU with $32$GB memory.
The hidden vector size $H$ and total layer number $L$ are both $1024$ and $24$, respectively, according to the PLMs adopted. 
We follow CSProm-KG to set the prompt length $n$ for a single transformed structural embedding as $10$, and the textual information for a triple to predict is truncated to a maximum of $72$ tokens. 
As for the disentangled graph learner, the dimension of structural component embedding and relation embedding is set to $200$. Correspondingly, the kernel sizes $k_w, k_h$ in ConvE become 10 and 20, respectively.
The number of the aggregation layer $L_{g}$ is set to 1.
The coefficient of Mutual Information regularization $\lambda$ is set to $0.1$.
We select the component numbers $K$ from a pool of $\{2, 4, 6\}$, and finally select an optimum one according to MRR on the validation set; it is $2$ for WN18RR, and $4$ for \revise{both} FB15K-237 \revise{and CoDEx-L}.
We implement our model with PyTorch and use Adam as optimizer with a learning rate of 0.0001 and a batch size of 64, which are the optimum configurations selected according to MRR on the validation set.

\subsection{Main Results}
The overall results are shown in Table \ref{tab:overall_results} \revise{and Table \ref{tab:Results_CoDEx}}.
As we can see, our models always outperform the baselines of \textit{Structure}-based, \textit{PLM}-based and \textit{Joint}. Especially, on FB15K-237, $\text{PDKGC}_{\text{BERT}}$[C] and $\text{PDKGC}_{\text{RoBERTa}}$[C] have achieved the top-2 best results,
$\text{PDKGC}_{\text{BERT}}$[S] outperforms all \textit{Structure}-based methods on most metrics, and $\text{PDKGC}_{\text{RoBERTa}}$[T] consistently beats all \textit{PLM}-based methods \revise{and outperforms 8 out of 9 methods} by a large margin. 
We also note that 1) the best [T] and [S] are achieved by different PLMs, indicating that different PLMs show different preferences in encoding the textual information, and 2) the gap between $\text{PDKGC}_{\text{BERT}}$[C] and $\text{PDKGC}_{\text{RoBERTa}}$[C] is slight, where our proposed ensemble solution is effective in combining the outputs of the two \textit{predictors}.
\revise{Compared to FB15K-237, CoDEx-L features equally rich relations (i.e., structural patterns) and more entities. While on CoDEx-L, as shown in Table \ref{tab:Results_CoDEx}, PDKGC with BERT achieves state-of-the-art performance against existing representative baselines, showing the effectiveness of PDKGC on larger KGs with diverse scopes and levels of difficulty.}


\revise{On WN18RR, t}he PDKGC variants achieve very competitive results.
When RoBERTa is used, PDKGC[C] has \revise{been the second best among all the methods,}
especially on MRR, Hits@3 and Hits@10. 
However, the outperformance on WN18RR is less promising in comparison with that on FB15K-237. This discrepancy may be due to that textually similar entities are more common in WN18RR. 
In contrast to our methods that solely rely on the given text and structural context to choose the missing entities, the baselines often raise \revise{extra} strategies to deal with this textual similarity.
For example, KG-S2S first collects all the tokens in the entity text to generate an Entity Prefix Trie and uses this prefix Trie to control the scope of decoding next token, leading to impressive scores on Hits@1. While CSProm-KG adds a LAR loss to filter textually similar entities, significantly narrowing down the space of possible correct entities. 
Therefore, we also report the results of $\text{CSProm-KG}_{\text{non-LAR}}$ with LAR removed, it can be seen that either $\text{PDKGC}_{\text{BERT}}$[C] or $\text{PDKGC}_{\text{RoBERTa}}$[C] outperforms it by a large margin on WN18RR.
\revise{Most notably, SKG-KGC not only introduces a large number of negative samples but also broadens the training set by packing the triples that share the same $(h,r)$ or $(r,t)$ into one new training triple, both of which enhance its ability to identity similar candidate entities, leading to the best results and significant outperformance over other methods on WN18RR.}
This also motivates us to try some effective solutions to tackle these textually similar entities.


\begin{table*}[htbp]
\centering
\caption{\small Prompt Analysis Results. PDKGC here uses BERT-large.
Among variants that have three prediction results, i.e., ``X[Y]'' and ``Y=\{T, S, C\}'', for each kind of prediction, we highlight the best ones using underline.
}
\label{tab:AB_deep_analysis}
\resizebox{0.8\linewidth}{!}{
\begin{tabular}{c|l|cccc|cccc}
\toprule[0.7pt]
\multicolumn{1}{c|}{\multirow{2}{*}{No.}}
& 
\multicolumn{1}{c|}{\multirow{2}{*}{Methods}}
&
\multicolumn{4}{c|}{WN18RR} & \multicolumn{4}{c}{FB15K-237}\\
& &  MRR &  Hits@1 & Hits@3 & Hits@10 &  MRR  & Hits@1 & Hits@3 &  Hits@10
\\
\midrule[0.5pt]
1 & DisenKGAT
& 0.486  & 0.441 & 0.502 & 0.578
& \textit{0.366} & \textit{0.271} & \textit{0.405} & \textit{0.553}
\\
\midrule[0.5pt]
2 & $\text{MEM-KGC}_{\text{fine-tuned}}$
& \it{0.530} & \it{0.483} & \it{0.559} & \it{0.611}
& \it{0.332}& \it{0.244} & \it{0.367} & \it{0.508}
\\\midrule[0.5pt]
3 &$\text{CSProm-KG}_{\text{non-LAR}}$
&0.534& 0.489& -- &0.624
& \textit{0.350} & \textit{0.259} & \textit{0.384}& 
\textit{0.530}
\\
\midrule[0.5pt]
4 & PDKGC [T]
& 0.531 & 0.450 & 0.571 & 0.688
& 0.348 & 0.260 & 0.383 & 0.526
\\
5 & PDKGC [S]
& \underline{0.543} & \underline{0.490} & \underline{0.565} & \underline{0.646}
& \underline{0.370} & \underline{0.278} & \underline{0.407} & \underline{0.551}
\\
6 & PDKGC [C]
& \underline{0.568} & \underline{0.500} & \underline{0.598} & \underline{0.702}
& \underline{0.381} & \underline{0.289} & \underline{0.418}& \underline{0.567}
\\
\midrule[0.5pt]
7 & $\text{PDKGC}_{\text{w/o Disen}}$[T]
&0.533&\underline{0.456}&0.571&0.678
&0.349&0.262&0.381&0.524
\\
8 & $\text{PDKGC}_{\text{w/o Disen}}$[S]
&0.536&0.490&0.551&0.627
&0.350&0.260&0.386&0.534
\\
9 & $\text{PDKGC}_{\text{w/o Disen}}$[C]
&0.557&0.500&0.577&0.669
&0.368&0.275&0.403&0.557
\\\midrule[0.3pt]
10 & $\text{PDKGC}_{\text{w/o TP}}$
&0.542&0.485&0.568&0.651
&0.366&0.276&0.402&0.545
\\\midrule[0.5pt]
11 & $\text{PDKGC}_{\text{single}}$[T]
& \underline{0.535} & 0.443 & \underline{0.573} & \underline{0.695} 
& \underline{0.354} & \underline{0.265} & \underline{0.387} & \underline{0.532}
\\
12 & $\text{PDKGC}_{\text{single}}$[S]
& 0.541 & 0.478 & 0.561 & 0.635
& 0.358 & 0.266 & 0.396 & 0.537
\\
13 & $\text{PDKGC}_{\text{single}}$[C]
& 0.566 & 0.489 & 0.591 & 0.701
& 0.378 & 0.283 & 0.417 & 0.567
\\
\bottomrule[0.7pt]
\end{tabular}
}
\end{table*}

There is another interesting observation from variants of MEM-KGC.
We can find that fine-tuning with BERT-large (i.e., $\text{MEM-KGC}_{\text{fine-tuned}}$) performs slightly worse than fine-tuning with BERT-base (i.e., the original MEM-KGC).
This is consistent with the observation of KG-BERT, and can be explained by the fact that BERT-base is simpler and less sensitive to hyper-parameter settings.
Meanwhile, when we freeze the parameters of BERT-large with only a simple linear classification layer fine-tuned (i.e., $\text{MEM-KGC}_{\text{frozen}}$), the performance dramatically drops, illustrating that this single linear layer is not ready for adapting the frozen PLMs to the KGC tasks.
In contrast, our PDKGC incorporates hard task prompts with disentangled structure embedding (prompts) to provide more valuable information for the frozen PLM to perform KGC and consequently has much better results.
We also notice that $\text{MEM-KGC}_{\text{frozen}}$ even performs better than some fine-tuned models such as KG-BERT, indicating the potential of frozen PLMs for inferring the missing triples.

\revise{We also observe that KGT5+Neighbors tends to be the worst on most metrics, it may be because the authors omit the entity descriptions and only keep the entity names for storing more neighbors in the input sequence, so that the model is hard to distinguish the given head entities. This also illustrates that KGT5+Neighbors is a naive idea that has much space to improve, while our PDKGC presents a more effective practice. }

\subsection{Impacts of the Prompts}
In this subsection, we use two PDKGC variants to analyze the impact of two critical techniques --- hard task prompt and disentangled structure prompt. 
One variant is $\text{PDKGC}_{\text{w/o Disen}}$, which replaces the disentangled entity embeddings with normal non-disentangled entity embeddings, for analysing the impact of the disentangled structure prompt (i.e., the disentangled graph learner module).
In $\text{PDKGC}_{\text{w/o Disen}}$, we directly generate the structural prompts based on the initialized entity embeddings without attentively aggregating features of different semantic aspects from the neighborhood.
The other variant is $\text{PDKGC}_{\text{w/o TP}}$, which removes the \textit{textual predictor} (TP) that is used to predict the missing entities through the hard task prompt given the triple context.
$\text{PDKGC}_{\text{w/o TP}}$ relies on the \textit{structural predictor} (SP) alone to make predictions.

We test these two variants on \revise{WN18RR and FB15K-237}, with the results reported in Table \ref{tab:AB_deep_analysis}.
It can be seen that their performance consistently drops in comparison with the original PDKGC.
More specifically, the outputs of $\text{PDKGC}_{\text{w/o Disen}}$'s \textit{structural predictor} and its ensemble results (Rows 8 and 9) are both inferior to those of PDKGC (Rows 5 and 6).
Besides, there are also some exceptions on the outputs of $\text{PDKGC}_{\text{w/o Disen}}$'s \textit{textual predictor} w.r.t. the metrics of MRR and Hits@1 (see Row 4 vs Row 7). 
This may be due to the weighted loss which aims to balance the prediction results of \textit{textual predictor} and  \textit{structural predictor}.
Meanwhile, their gaps are relatively small.
Through comparing Row 10 with Rows 5 and 6, we can observe that $\text{PDKGC}_{\text{w/o TP}}$ performs worse than PDKGC, sometimes even worse than PDKGC's \textit{structural predictor}.
Therefore, we can conclude that both disentangled structure prompt and hard task prompt are contributory modules of PDKGC.

In comparison with $\text{CSProm-KG}_{\text{non-LAR}}$, the difference of $\text{PDKGC}_{\text{w/o Disen}}$ mainly lies in an additional TP module; in comparison with MEM-KGC, $\text{PDKGC}_{\text{w/o Disen}}$ additionally utilizes structural information. 
As shown in Table \ref{tab:AB_deep_analysis}, $\text{PDKGC}_{\text{w/o Disen}}$ performs better than both $\text{CSProm-KG}_{\text{non-LAR}}$ and MEM-KGC.
Also, compared with $\text{CSProm-KG}_{\text{non-LAR}}$, $\text{PDKGC}_{\text{w/o TP}}$ additionally concerns the disentangled entity representations; compared with DisenKGAT, $\text{PDKGC}_{\text{w/o TP}}$ additionally considers textual information.
Similarly, from Table \ref{tab:AB_deep_analysis}, $\text{PDKGC}_{\text{w/o TP}}$ is superior to both $\text{CSProm-KG}_{\text{non-LAR}}$ and DisenKGAT on most metrics.
As a result, we can say our method outperforms $\text{CSProm-KG}_{\text{non-LAR}}$ through the prediction from the text side by hard task prompt and the disentangled structural prompt, and outperforms MEM-KGC and DisenKGAT through effectively adding the structural information and textual information, respectively.
All these further verify the effectiveness of the disentangled structure prompt and the hard task prompt we proposed.

\begin{figure}[htbp]
  \centering  \includegraphics[width=0.95\linewidth]{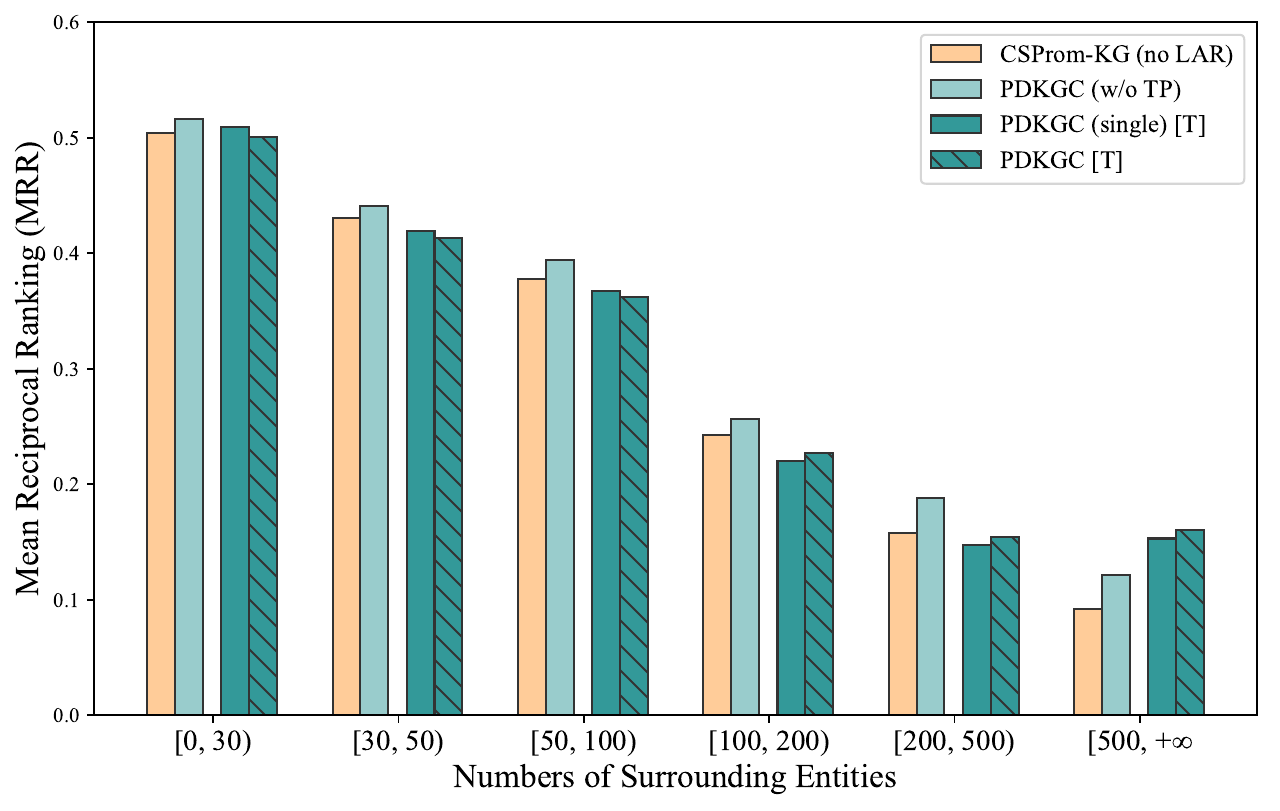}
\caption{\small Performance (MRR) of different models on different testing triples of FB15K-237, which have different numbers of surrounding entities. ``PDKGC [T]'' and ``PDKGC (single) [T]'' means the results from PDKGC and $\text{PDKGC}_{\text{single}}$'s \textit{textual predictors}, respectively.}
\label{fig:AB_degree_count}
\end{figure}

We further have quantitative results to analyze the discrepancy between $\text{PDKGC}_{\text{w/o TP}}$ and $\text{CSProm-KG}_{\text{non-LAR}}$, they differ in taking into account the complex neighborhood environment around a triple to complete, i.e., its surrounding entities.
Therefore, for all the triples in the testing set, we group them according to the range of the number of entities surrounding the known entities (i.e., the given head (resp. tail) entities for \textit{tail (resp. head) entity prediction}), and re-calculate the result of MRR of each group.
The results on FB15K-237 are shown in Fig. \ref{fig:AB_degree_count}, from which we can see that the performance gap between $\text{PDKGC}_{\text{w/o TP}}$ and $\text{CSProm-KG}_{\text{non-LAR}}$ is enlarged as the number of surrounding entities increases.
Statistically, the relative performance gain from $\text{CSProm-KG}_{\text{non-LAR}}$ to $\text{PDKGC}_{\text{w/o TP}}$ is $2.35\%$, $2.4\%$, $4.2\%$, $5.8\%$, $19.3\%$, $32.5\%$ across the six increasing intervals.
This indicates that our proposed disentangled structure prompt is more effective in processing the structural information, especially when the neighborhood is complex with semantics of different aspects.







\subsection{Ablation Studies}
\vspace{0.15cm}
\subsubsection{Impact of using all or one of the disentangled components}\label{sec:ab_single_component}
As introduced in Section \ref{sec:structure_aware_text_encoder}, for a triple e.g. $(h,r,?)$ to complete, we feed all the disentangled component embeddings of $h$ into the PLM instead of selecting one component that is highly relevant to the triple.
To illustrate the effectiveness of this choice, we implement another model variant $\text{PDKGC}_{\text{single}}$ for comparison, which first assesses all the components by measuring their relatedness with the relation in the triple, and then feeds the most relevant component into the PLM as the structure prompt.
The relatedness is measured by computing the dot-product similarity of two structural embeddings.

The results are shown in the last three rows of Table \ref{tab:AB_deep_analysis}.
We can see that $\text{PDKGC}_{\text{single}}$ performs worse than PDKGC w.r.t the \textit{structural predictor} (Row 5 vs 12) and the ensemble (Row 6 vs 13), but performs a bit better w.r.t. the \textit{textual predictor} (Row 4 vs 11).
One potential reason is that using the selected single component loses some information in comparison with using all the components, especially when the neighborhood of the triple is complex, leading to worse performance of the \textit{structural predictor} and the ensemble.
However, it sometimes could provide useful and more compact structural information to the \textit{textual predictor}, leading to a bit better performance.
To further verify this, we calculate the results of the \textit{textual predictors} of PDKGC and $\text{PDKGC}_{\text{single}}$ on different testing triple sets with different numbers of neighboring entities, as shown in Fig. \ref{fig:AB_degree_count}.
It can be seen that PDKGC performs better when the number of neighboring entities exceeds 100, showing the superiority of using all the components when the neighborhood is complex. 
We have more observations about this in Section \ref{sec:case_study}.

\subsubsection{
Impact of KGE models}
As we have mentioned, our PDKGC is flexible to incorporate different KGE models.
To validate this, we replace the applied ConvE with another two popular KGE models: TransE and DistMult, and conduct evaluations with BERT-large.
We report the ensemble results of PDKGC on WN18RR in Table \ref{tab:KGE_results}.
As we can see, PDKGC successfully cooperates with these KGE models and significantly outperforms their original versions.
Moreover, we also list the public results of CSProm-KG when it incorporates these KGE models.
It can be seen that with TransE and DistMult, PDKGC achieves higher performance gains compared with CSProm-KG in most situations, illustrating that our PDKGC is more robust when shifting to other KGE models, while the good performance of CSProm-KG with ConvE may be partially due to the careful selection of hyper-parameters, especially those for optimizing the LAR module.


\begin{table}[htbp]
\centering
\caption{\small The ensemble results of  $\text{PDKGC}_{\text{BERT}}$[C] on WN18RR when different KGE models are applied.
}
\label{tab:KGE_results}
\resizebox{\linewidth}{!}{
\begin{tabular}{p{1.7cm}|p{1.4cm}<{\centering}p{1.4cm}<{\centering}p{1.4cm}<{\centering}p{1.4cm}<{\centering}}
\toprule[0.7pt]
 \multicolumn{1}{c|}{Methods} &  MRR &  Hits@1 & Hits@3 & Hits@10 
\\
\midrule[0.5pt]
\footnotesize{TransE}
&0.243 & 0.043 & 0.441 &0.532
\\
\  \scriptsize{+ CSProm-KG} 
& 0.499\scriptsize{\textcolor{blue}{($\uparrow$0.256)}}
& 0.462\scriptsize{\textcolor{blue}{($\uparrow$0.419)}} 
& 0.515\scriptsize{\textcolor{blue}{($\uparrow$0.074)}}
& 0.569\scriptsize{\textcolor{blue}{($\uparrow$0.037)}}
\\
\  \scriptsize{+ PDKGC}
&0.542\scriptsize{\textcolor{red}{($\uparrow$0.299)} }
&0.466\scriptsize{\textcolor{red}{($\uparrow$0.423)}}
&0.575\scriptsize{\textcolor{red}{($\uparrow$0.134)}}
&0.690\scriptsize{\textcolor{red}{($\uparrow$0.158)}}
\\\midrule[0.5pt]
\footnotesize{DistMult}
& 0.444  & 0.412 & 0.470 & 0.504
\\
\  \scriptsize{+ CSProm-KG}
& 0.543\scriptsize{\textcolor{blue}{($\uparrow$0.099)}}
& 0.494\scriptsize{\textcolor{blue}{($\uparrow$0.082)}}
& 0.562\scriptsize{\textcolor{blue}{($\uparrow$0.092)}}
& 0.639\scriptsize{\textcolor{blue}{($\uparrow$0.135)}}
\\
\  \scriptsize{+ PDKGC}
&0.560\scriptsize{\textcolor{red}{($\uparrow$0.116)}}
&0.491\scriptsize{\textcolor{red}{($\uparrow$0.079)}}
&0.588\scriptsize{\textcolor{red}{($\uparrow$0.118)}}
&0.693\scriptsize{\textcolor{red}{($\uparrow$0.189)}}
\\\midrule[0.5pt]
\footnotesize{ConvE}
 & 0.456  & 0.419 & 0.470 & 0.531 
\\
\ \scriptsize{+ CSProm-KG}
&0.575\scriptsize{\textcolor{blue}{($\uparrow$0.119)} }
&0.522\scriptsize{\textcolor{blue}{($\uparrow$0.103)}}
&0.596\scriptsize{\textcolor{blue}{($\uparrow$0.126)}}
&0.678\scriptsize{\textcolor{blue}{($\uparrow$0.147)}}
\\
\  \scriptsize{+ PDKGC}
& 0.568\scriptsize{\textcolor{red}{($\uparrow$0.112)}}
& 0.500\scriptsize{\textcolor{red}{($\uparrow$0.081)}}
& 0.598\scriptsize{\textcolor{red}{($\uparrow$0.128)}}
& 0.702\scriptsize{\textcolor{red}{($\uparrow$0.171)}}
\\
\bottomrule[0.7pt]
\end{tabular}
}
\end{table}


\begin{figure*}[htbp]
  \centering
  \includegraphics[width=0.95\linewidth]{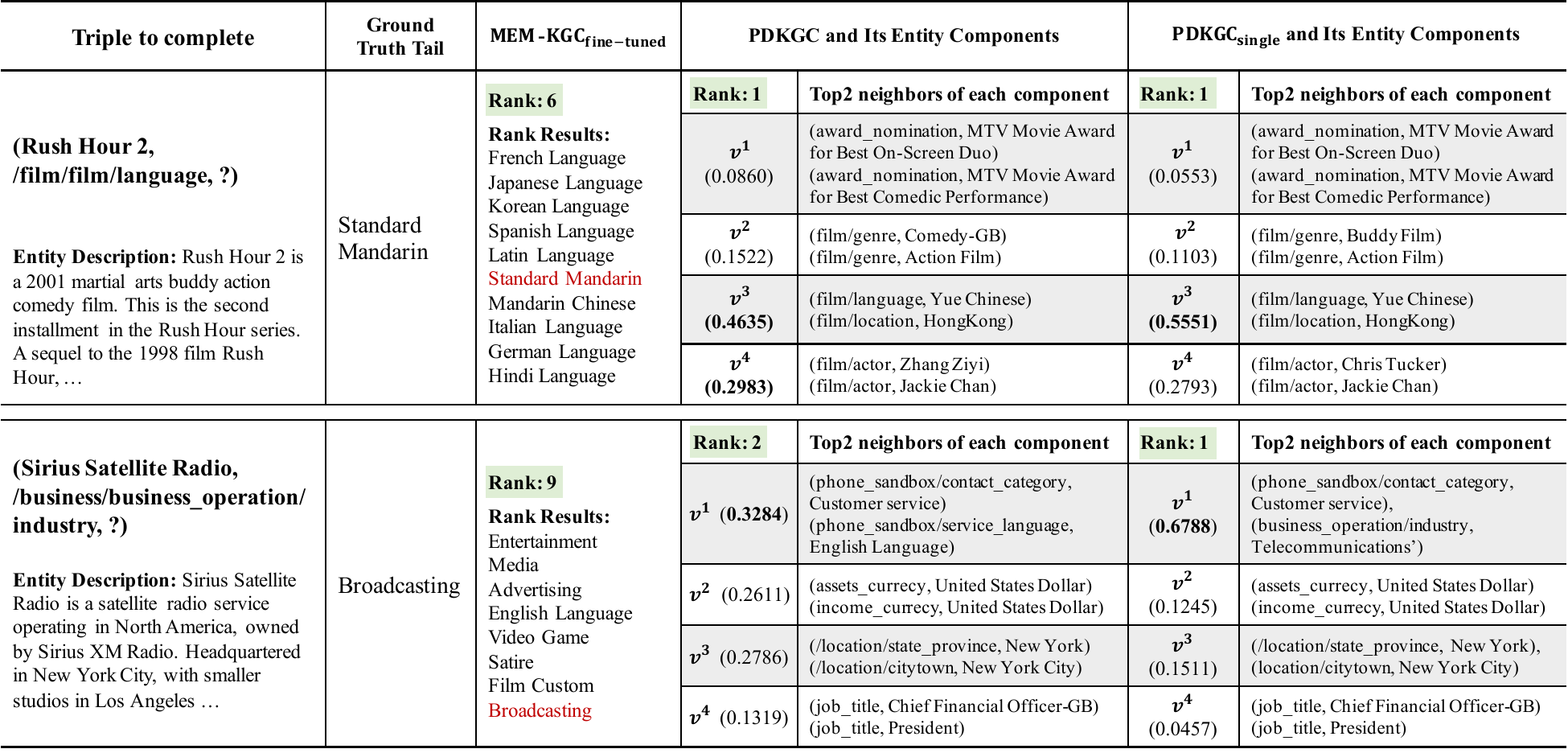}
\caption{\small Two triples to complete from FB15K-237's testing set. The ranks predicted by different models are highlighted using green background. For PDKGC and $\text{PDKGC}_{\text{single}}$, we report the ranks produced by their \textit{textual predictors}, and also present the disentangled components of the head entity including the correlation weight between the component and the target triple, and the component's corresponding Top-2 neighbors.}
\label{fig:case_study}
\end{figure*}

\subsection{Case Studies}\label{sec:case_study}
In Fig. \ref{fig:case_study}, we present two testing triples to complete, with the descriptions of their head entities, their ground truth tail entities, and the entity ranking results by different models.
For PDKGC and $\text{PDKGC}_{\text{single}}$, we not only present the ranking result but also demonstrate the head entity's disentangled embedding and neighbors.
More specifically, for each entity, we leverage the attention value $\alpha$ to indicate its correlations to different neighbors w.r.t each component of the disentangled embeddings.
For example, in the first testing triple of Fig. \ref{fig:case_study}, the top-2 most similar neighbors of head entity \textit{Rush Hour 2} w.r.t. the second component $\bm{v}^2$ are (\textit{film/genre, Comedy-GB})\footnote{The complete triple is (\textit{Rush Hour 2, /film/film/genre, Comedy-GB}). As shown above, the relation is abbreviated, and each neighbor is represented by the format of (relation, tail) with reversed relations concerned.} and (\textit{film/genre, Action Film}), while (\textit{film/actor, Zhang Ziyi}) and (\textit{film/actor, Jackie Chan}) are the top-2 w.r.t. the fourth component $\bm{v}^4$.
It is clear that different components are related to different neighbors, representing different semantic aspects of the head entity.
Also, we can use these neighbors to explain the semantic meanings of each component.
For example, we assume that the four components of \textit{Rush Hour 2} separately reflect its four semantic aspects: awards received, category, film features, and actor\&actress.

Moreover, for each testing triple, we also compute a dot-product similarity between the structural embeddings of relation and entity component to highlight the relevant components for the current inference. As shown in the first case of Fig. \ref{fig:case_study}, with PDKGC, the components $\bm{v}^3$ and $\bm{v}^4$ of \textit{Rush Hour 2}, which contain the information about film language\&location features and film actors\&actresses, respectively, contribute more effective clues to infer the tail entity, while the other components play a much less important role.
We can use these component weights to provide an explanation for each prediction, justifying whether the prediction is good or bad.

Back to the rank results produced by different models, it can be observed that our PDKGC and $\text{PDKGC}_{\text{single}}$ both have higher ranks than $\text{MEM-KGC}_{\text{fine-tuned}}$ (a version fine-tuned using BERT-large) over two examples.
For example, in the second case of Fig. \ref{fig:case_study}, the rank of the ground truth tail \textit{Broadcasting} is raised from $9$ to $2$ by PDKGC, and to $1$ by $\text{PDKGC}_{\text{single}}$.
This indicates that our model variants successfully provide discriminative structural information except for the textual descriptions of the head entity and relation given in the target triple.
Moreover, according to the disentangled components highlighted by our models, we find that PDKGC and $\text{PDKGC}_{\text{single}}$ both extract useful structural information from the neighborhood, i.e., the neighbors (\textit{phone\_sandbox/contact\_category, Customer service}) and (\textit{business\_operation/industry, Telecommunications}) both indicate that \textit{Sirius Satellite Radio} is more likely to be a Broadcasting company.

From the second example in Fig. \ref{fig:case_study}, we also find that the component $\bm{v}^1$ learned by $\text{PDKGC}_{\text{single}}$ contains more accurate information than that learned by PDKGC, this is in line with our observations in Section \ref{sec:ab_single_component}, where $\text{PDKGC}_{\text{single}}$ may gather less irrelevant structure semantics by feeding only a single component into PLMs.
However, the number of neighbors around \textit{Sirius Satellite Radio} is limited to 16.
When we face the first example in Fig. \ref{fig:case_study} where the number of neighbors around \textit{Rush Hour 2} is much more, PDKGC also highlights component $\bm{v}^4$ to predict the tail with higher confidence.
Namely, besides the facts implied in $\bm{v}^3$ (i.e., the film location is \textit{Hong Kong} and the film language is \textit{Yue Language}), PDKGC can more accurately predict the tail as \textit{Standard Mandarin} according to the facts that the film has two Chinese famous actors/actresses \textit{Jackie Chan} and \textit{Zhang Ziyi} implied in $\bm{v}^4$.
This observation verifies that PDKGC keeps more complete structural information than $\text{PDKGC}_{\text{single}}$ for each triple to complete.

\subsection{Model Efficiency}
In comparison with many \textit{PLM}-based and joint KGC methods, our PDKGC has higher training and inference efficiency.
In this part, we will give theoretical as well as practical proof on this point.

\subsubsection{Theoretical Analysis}
As the computation overheads are mainly happening inside the PLMs, we focus on computing the complexity of performing the (multi-head) self-attention in a Transformer block, which is $\mathcal{O}({L_{s}}^2 H)$ with $L_{s}$ as the length of the input sequence and $H$ as the embedding size.
Here, we use $\mathcal{O}({L_{s}}^2)$ to simply represent the complexity as $H \ll {L_{s}}^2$.

\textbf{Training Complexity.}
We first analyze the time complexity of different models w.r.t. one training triple. 
When methods such as KG-BERT and LASS forward the full text of a triple to the PLM with a sequence length of $L_s$, their time complexity is thus illustrated as ${L_s}^2$.
In practice, the sequence lengths of the two split parts in a triple, i.e., $(h,r$) and $t$, or $h$ and $(r,t)$, are similar because the length of an entity's text is usually longer than a relation's text, especially when the entity description is included.
Therefore, for methods that separately process the two different parts of a triple by Siamese PLMs, e.g., StAR and SimKGC, or encoder-decoder PLMs, e.g., KG-S2S \revise{and KGT5}, the complexity is computed as $(L_s/2)^2 + (L_s/2)^2$.
\revise{Similarly, the complexity of SKG-KGC and KGT5+Neighbors is computed as $(L'_s/2)^2 + (L_s/2)^2$ with $L'_s \textgreater L_s$, where $L'_s$ in KGT5+Neighbors means the length of the sequence that contains $h$ or $t$'s serialized neighborhood information, while $L'_s$ in SKG-KGC means the length of the sequence that represents more than one entity, i.e., entity set that associated with the same $(h,r)$ or $(r,t)$.}
As for MEM-KGC, CSPromKG and our PDKGC, only encoding $(h,r)$ or $(r,t)$ is required during training, the corresponding complexity is $(L_s/2)^2$.
Note the applied prompt is far shorter than the entity description.

\begin{table}
\centering
\caption{\small Comparisons of different models w.r.t. their training and inference efficiency.}
\begin{subtable}{1\linewidth}
\centering
\caption{\small Inference Complexity. $L_s$ here means the length (token number) of the text of a triple. $BS$ is for \textit{beam search} and computed as $|V| \times m \times L_s/2$ where $m$ is its beam size and $|V|$ is the applied PLM's vocabulary size. $N_{te}$ denotes the number of all the test triples.}
\label{tab:test_complexity}
\resizebox{\linewidth}{!}{
\begin{tabular}{c|ccc|ccc}
\toprule[0.7pt]
Model & One Triple & All Triples \\
\midrule[0.5pt]
KG-BERT, LASS 
& ${L_s}^2 \times |\mathcal{E}|$
& ${L_s}^2 \times |\mathcal{E}| \times N_{te}$
\\\midrule[0.5pt]
StAR, SimKGC, \revise{SKG-KGC}
& $(L_s/2)^2 \times (1+|\mathcal{E}|)$
& $(L_s/2)^2 \times (N_{te}+|\mathcal{E}|)$
\\\midrule[0.5pt]
KG-S2S, \revise{KGT5}
& $(L_s/2)^2 + BS$ 
& $((L_s/2)^2 + BS) \times N_{te}$
\\\midrule[0.5pt]
\makecell{CSProm-KG, MEM-KGC, \textbf{PDKGC}}
& $(L_s/2)^2$
& $(L_s/2)^2 \times N_{te}$
\\\bottomrule[0.7pt]
\end{tabular}
}
\end{subtable}%
\newline
\newline
\begin{subtable}{1\linewidth}
\centering
\caption{\small Training and Inference time comparison on WN18RR. \#Total and \#Trainable denote the total and trainable parameters. T/Ep and Inf denote the training time per epoch and inference time.}
\label{tab:train_infer_time}
\resizebox{0.95\linewidth}{!}{
\begin{tabular}{c|ccccc}
\toprule[0.7pt]
Model & PLM & \# Total & \# Trainable & T/Ep & Inf 
 \\
\midrule[0.5pt]
StAR & RoBERTa-large & 355M & 355M & 75m & 23m
\\\midrule[0.5pt]
\multirow{2}{*}{KG-S2S} 
&  T5-base & 222M & 222M & 7m & 17m
\\
& T5-large & 737M & 737M & 16m & 23m
\\\midrule[0.5pt]
\multirow{2}{*}{CSProm-KG }  & BERT-base & 126M  & 17M & 3m & 6s
\\
 & BERT-large & 363M & 28M & 7m & 10s
\\\midrule[0.5pt]
\multirow{3}{*}{PDKGC (ours)} 
& BERT-base & 127M & 18M & 5m & 10s
\\
& BERT-large & 365M & 30M & 10m & 13s
\\
& RoBERTa-large & 385M & 30M & 10m & 13s
\\
\bottomrule[0.7pt]
\end{tabular}
}
\end{subtable}%
\end{table}

Besides the complexity of encoding each triple, some methods raise negative training triples, which require additional computation.
SimKGC \revise{and SKG-KGC take} entities in the same batch and entities in one or two previous batches as negatives, \revise{where the batch size is by default set to 1024}.
StAR follows KG-BERT to set 5 corrupted triples for one positive.
In contrast, our method PDKGC requires no negative samples during the training.

\textbf{Inference Complexity.}
As shown in Table \ref{tab:test_complexity}, we list the inference complexities in two cases, i.e., only one test triple and all the test triples.
As we have introduced more than once, the combinatorial explosion in the testing stage of KG-BERT and LASS often results in huge time cost, where all the entities in the dataset are taken as candidates to send into the BERT for predicting each triple.
StAR, SimKGC \revise{and SKG-KGC} require well-trained PLM outputs to represent all entities at first during testing.
KG-S2S struggles in decoding the textual descriptions of the missing entities token by token with the costly beam search.
\revise{KGT5+Neighbors performs similarly but has longer input sequences due to serialized neighboring entities and relations.}
Our method PDKGC as well as CSProm-KG and MEM-KGC are expected to be the fastest because they infer scores of all entities at once, by comparing entity representations with the encoded representation of the other two known elements in a triple or sending the the encoded representation into the entity classifier.

\subsubsection{Practical Verification}
Since different methods adopt different PLMs with various parameter sizes, as well as methods like CSProm-KG and our PDKGC introduce additional modules for learning prompts, we also run them on the same device (i.e., a single Tesla V100 GPU) and compare their actual training and inference time.
It is noted that we omit the comparisons with KG-BERT and LASS since they are theoretically slower in training and testing. Besides, we also select one representative method among those that have similar time complexity. 
The results are shown in Table~\ref{tab:train_infer_time}.

It can be seen that \textit{i)} CSProm-KG and our PDKGC take the least amount of time in both training and testing when applying the same scale of PLMs, which is consistent with our conclusions raised in the theoretical analysis part; and \textit{ii)} PDKGC is slightly slower than CSProm-KG, this is because our PDKGC includes a GNN module for aggregating the neighborhood features around each entity and learning the disentangled structure prompt, which introduces some computational cost but achieves higher KGC performance. However, it is still faster in training and much faster in testing than KG-S2S and StAR.
We also notice that compared with KG-S2S, StAR which has the same training time complexity takes more time in training especially when KG-S2S uses the large version of T5; this is mainly due to the additional negative training samples used in StAR.

Another advantage of CSProm-KG and PDKGC in training efficiency is their fewer trainable parameters, as shown in Table \ref{tab:train_infer_time}.
This smaller number of tunable parameters enables them less sensitive to the hyperparameter choices.



\section{Conclusion and Outlook}
In this paper, we proposed PDKGC, a novel prompt-tuning-based method for KG completion, which is built upon one frozen pre-trained language model (PLM) and two different prompts for fully utilizing the text encoding knowledge learned in the PLM and effectively incorporating structural knowledge with disentangled KG embeddings.
Briefly, PDKGC includes \textit{(i)} a disentangled graph learner with the relation-aware attention mechanism applied to distribute the neighbors on the graph to learn different representation components, each of which encodes one independent aspect of the structure semantics; \textit{(ii)} a \textit{textual predictor} which translates the graph representation components into a disentangled structure prompt and feeds it to the frozen PLM together with a hard task prompt from the triple text to predict the missing entity;
\textit{(iii)} a \textit{structural predictor} which feeds the PLM outputs of these structural prompts into a KGE model for entity prediction.
The textual and structural predictors complement each other, and their ensemble leads to better performance.
In evaluation, we conducted solid experiments on \revise{three} widely-used KGC benchmarks, and compared PDKGC with traditional \textit{structure}-based methods as well as state-of-the-art fine-tuned and frozen \textit{PLM}-based methods.
PDKGC often achieves better performance, and the effectiveness of the two proposed prompts has been fully verified. 
In the future, we will consider encoder-decoder and decoder-only PLMs with a larger size, and extend PDKGC to inductive KGC which aims to complete triples with new entities and/or relations that have never appeared during training \citep{geng2023relational}.

\section*{Acknowledgements}
\revise{This work is funded in part by the Zhejiang Provincial Natural Science Foundation of China under Grant No. LQ24F020034 and No. Q23F020051, the Natural Science Foundation of China under Grant No. 62072149 and No. 62306276, the Primary R\&D Plan of Zhejiang under Grant No. 2023C03198.
Jiaoyan Chen is mainly supported by the EPSRC project OntoEm (EP/Y017706/1), Jeff Z. Pan is mainly supported by the Chang Jiang Scholars Program (J2019032).} 


\bibliographystyle{cas-model2-names}

\bibliography{ref.bib}



\end{document}